\documentclass[journal,twoside,web]{ieeecolor}
\usepackage{tmi}
\usepackage{cite}
\usepackage{amsmath,amssymb,amsfonts}
\usepackage{algorithm}
\usepackage{algpseudocode}
\usepackage{graphicx}
\usepackage{textcomp}
\usepackage{algorithm}
\usepackage{algpseudocode}

\algrenewcommand\algorithmicrequire{\textbf{Require:}}
\algrenewcommand\algorithmicensure{\textbf{Ensure:}}

\makeatletter
\renewcommand{\ALG@name}{Algorithm}
\makeatother

\usepackage{booktabs}
\usepackage{multirow}
\usepackage{threeparttable}
\def\BibTeX{{\rm B\kern-.05em{\sc i\kern-.025em b}\kern-.08em
    T\kern-.1667em\lower.7ex\hbox{E}\kern-.125emX}}
\usepackage{placeins}
\begin{document}
\title{Back to Physics: Operator-Guided Generative Paths for SMS MRI Reconstruction}
\author{
Zhibo~Chen,
Yu~Guan,
Yajuan~Huang,
Chaoqi~Chen,
Xiang~Ji,
Qiuyun~Fan,
\\Dong~Liang,~\IEEEmembership{Senior Member,~IEEE},
and Qiegen~Liu,~\IEEEmembership{Senior Member,~IEEE}%
\thanks{This work was supported in part by the National Key Research and Development Program of China under Grant 2023YFF1204300 and Grant 2023YFF1204302.  ( Zhibo Chen and Yu Guan are co-first authors.) (Corresponding authors: Dong Liang; Qiegen Liu.)}
\thanks{Zhibo Chen, Yajuan Huang, Xiang Ji, and Qiegen Liu are with the School of Information Engipeering, Nanchang University, Nanchang 330031, China(e-mail: 416100240222@email.ncu.edu.cn; guanyu@ncu.edu.cn; 6105123171@email.ncu.edu.cn; tojixiang@ncu.edu.cn; liuqiegen@ncu.edu.cn).}
\thanks{ Yu Guan is with School of Advanced Manufacturing, Nanchang University, Nanchang, 330031,China(e-mail: guanyu@ncu.edu.cn).}
\thanks{Chaoqi Chen is with the College of Computer Science and Software Engineering, Shenzhen University, Shenzhen 518060, China (e-mail: cqchen1994@gmail.com).}
\thanks{Qiuyun Fan is with the Academy of Medical Engineering and Translational01Medicine, Medical School, Faculty of Medicine, Tianjin University, Tianjin300072, China (e-mail: fanqiuyun@tju.edu.cn).}
\thanks{Dong Liang is with the Lauterbur Research Center for Biomedical Imagingand the Research Center for Medical AI, Shenzhen Institute of AdvancedTechnology, Chinese Academy of Sciences, Shenzhen 518055, China (e-mail: dong.liang@siat.ac.cn).}
}

\maketitle

\begin{abstract}
Simultaneous multi-slice (SMS) imaging with in-plane undersampling enables highly accelerated MRI but yields a strongly coupled inverse problem with deterministic inter-slice interference and missing k-space data.
Most diffusion-based reconstructions are formulated around Gaussian-noise corruption and rely on additional consistency steps to incorporate SMS physics, which can be mismatched to the operator-governed degradations in SMS acquisition.
We propose an operator-guided framework that models the degradation trajectory using known acquisition operators and inverts this process via deterministic updates.
Within this framework, we introduce an operator-conditional dual-stream interaction
network (OCDI-Net) that explicitly disentangles target-slice content from inter-slice
interference and predicts structured degradations for operator-aligned inversion,
and we instantiate reconstruction as a two-stage chained inference procedure that
performs SMS slice separation followed by in-plane completion.
Experiments on fastMRI brain data and prospectively acquired in vivo diffusion MRI data demonstrate improved fidelity and reduced slice leakage over conventional and learning-based SMS reconstructions.
\end{abstract}

\begin{IEEEkeywords}
Simultaneous multi-slice MRI, k-space reconstruction, operator-guided modeling, physics-informed learning
\end{IEEEkeywords}

\section{Introduction}
\label{sec:introduction}

\IEEEPARstart{S}{imultaneous} multi-slice (SMS) imaging accelerates MRI acquisition by exciting multiple slices within a single readout~\cite{r4,r5,r6,r9}.
Controlled Aliasing in Parallel Imaging (CAIPI) \cite{r1} applies slice-specific phase modulation to shape the aliasing pattern, which can be interpreted as controlled field-of-view (FOV) shifts along the phase-encoding direction. As a result, the acquired k-space measurement is not a noisy version of a single slice, but a deterministic superposition of multiple slices governed by known acquisition operators.
In practice, SMS is often combined with in-plane undersampling to reduce phase-encoding steps, which is particularly beneficial for high-resolution and time-sensitive scans such as diffusion-weighted imaging (DWI).
However, high multiband (MB) factors together with in-plane acceleration introduce a strongly ill-posed reconstruction problem, particularly under high acceleration, that is prone to slice leakage \cite{r7,r8} and loss of fine structural details.

\begin{figure}[!t]
  \centering
  \includegraphics[width=0.48\textwidth]{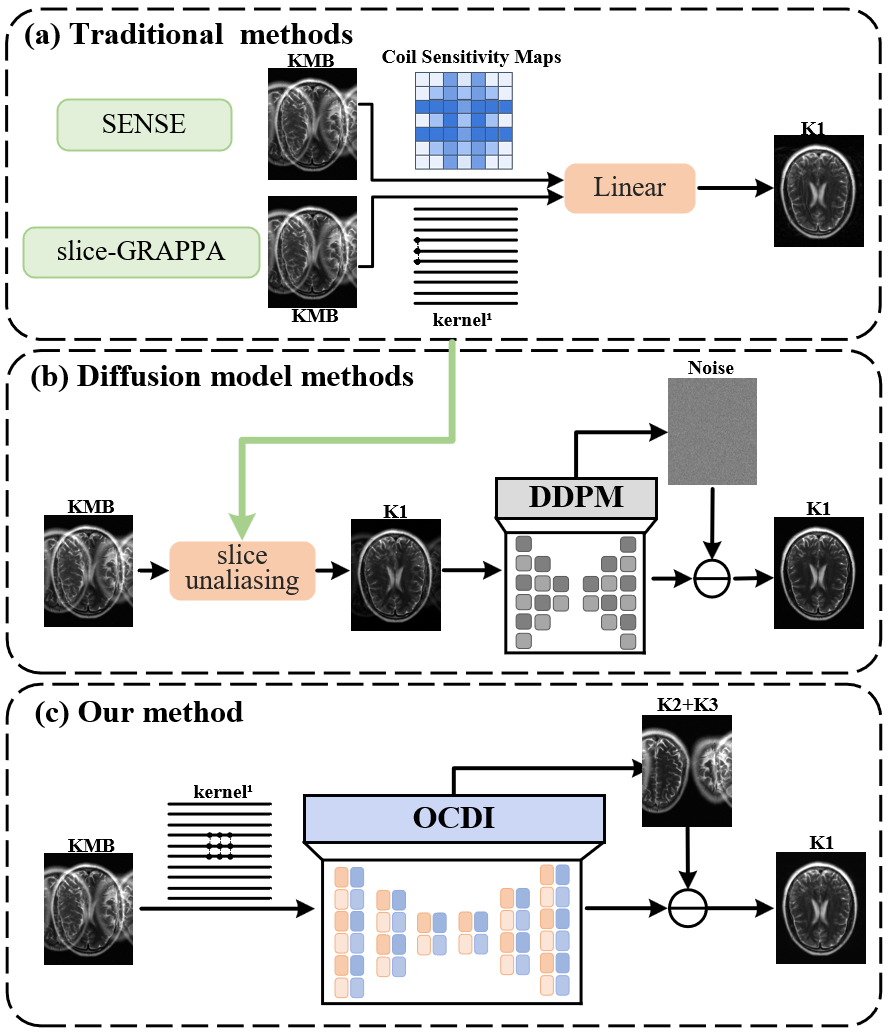}
  \vspace{-3mm}
  \caption{Conceptual comparison of SMS reconstruction paradigms.
(a) Linear operator-based methods recover slices from collapsed measurements.
(b) Diffusion-based approaches combine unaliasing with stochastic denoising and iterative refinement.
(c) Our operator-guided method predicts operator-induced components for deterministic inversion.
\emph{KMB} is the collapsed measurement, \emph{K1} is the target slice, and \emph{K2+K3} denotes the predicted interference removed to recover \emph{K1}.}
  \label{fig:framework_comparison}
\end{figure}

Despite their different characteristics, both degradations in SMS MRI originate from deterministic acquisition processes.
CAIPI-modulated slice superposition produces repeatable and structured interference patterns across slices, while Cartesian in-plane undersampling deterministically removes k-space information according to a predefined sampling mask.
Traditional SMS reconstruction methods, including SENSE \cite{r2} and GRAPPA-family \cite{r3} approaches, explicitly model these acquisition operators and recover individual slices via linear unmixing of the collapsed measurement through coil-sensitivity encoding or calibration-based kernel interpolation, as conceptually summarized in Fig.~\ref{fig:framework_comparison}(a).
Although physically grounded, these linear and local operators have limited capacity to recover high-frequency details under strong coupling between SMS folding and in-plane undersampling.
Recent diffusion-based reconstructions improve prior modeling capacity. However, most existing diffusion-based reconstruction methods are formulated around stochastic Gaussian noise corruption and rely on external data-consistency projections to enforce SMS physics during inference, as summarized in Fig.~\ref{fig:framework_comparison}(b).
This creates a common mismatch: Standard diffusion models assume a random walk guided by Gaussian perturbations, whereas the SMS degradation follows a directed trajectory governed by physical operators.
Consequently, they can leave coherent leakage patterns or oversmooth high-frequency structures under stronger SMS and in-plane acceleration.

In this work, we address the above mismatch by rethinking the generative reconstruction process from the perspective of acquisition physics.
Instead of modeling SMS reconstruction as the removal of stochastic noise, we treat both inter-slice interference and in-plane k-space incompleteness as deterministic degradations induced by known acquisition operators.
Under this view, the generative inference trajectory should be explicitly guided by the same operators that govern data acquisition, so that each intermediate state remains physically meaningful.
This operator-aligned formulation allows the reconstruction process to progressively explain the observed measurement in terms of structured, repeatable interference patterns and missing-data corruption, rather than absorbing them into generic noise models.

Building upon this principle, we propose an operator-guided k-space reconstruction framework that instantiates the above operator-aligned generative trajectory via a learned degradation predictor, as illustrated in Fig.~\ref{fig:framework_comparison}(c).
Specifically, we introduce an operator-conditional dual-stream interaction network (OCDI-Net) as a practical realization of this formulation, and adopt a staged inference strategy to decouple the strongly coupled effects of SMS slice superposition and in-plane undersampling.
This design improves stability under high acceleration and enables consistent performance gains as the multiband factor and in-plane acceleration increase.

We evaluate the proposed method on retrospectively simulated fastMRI brain data and prospectively acquired in vivo DWI data.
The results demonstrate improved reconstruction fidelity and reduced slice leakage compared with conventional and representative learning-based SMS reconstruction baselines.

The main contributions of this work are summarized as follows:
\begin{itemize}
  \item We propose an operator-guided k-space generative framework for SMS MRI by replacing Gaussian-noise corruption with an operator-induced deterministic degradation path and performing operator-aligned inversion.

  \item We introduce OCDI-Net, an operator-conditional dual-stream interaction network that disentangles target content from structured inter-slice interference and predicts operator-induced degradations for operator-aligned inversion, together with a two-stage inference strategy that decouples SMS slice separation and in-plane completion.

  \item We further demonstrate, through extensive experiments on fastMRI brain data and prospectively acquired DWI data, improved reconstruction fidelity and reduced slice leakage compared with conventional and representative learning-based SMS baselines.
\end{itemize}
\FloatBarrier

\section{Related Works}
\label{sec:related_work}

\subsection{Problem Formulation}

Simultaneous multi-slice acquisition excites multiple slices within a single readout, resulting in a slice-dependent phase-modulated superposition in the received multi-coil signal.
Let $\mathbf{k}_s \in \mathbb{C}^{N_x \times N_y \times N_c}$ denote the fully sampled multi-coil k-space of slice $s$, where $N_c$ is the number of receiver coils.
With multiband factor $B$, CAIPI-type encoding applies slice-specific phase ramps (equivalently, controlled FOV shifts in the image domain) prior to summation.
The collapsed multiband k-space can be expressed as
\begin{equation}
\mathbf{k}_{\mathrm{MB}} = \sum_{s=1}^{B} \mathcal{C}_s\!\left(\mathbf{k}_s\right),
\label{eq:rw_sms_superposition}
\end{equation}
where $\mathcal{C}_s(\cdot)$ denotes the CAIPI modulation operator for slice $s$.

In practice, SMS acquisition is often combined with in-plane undersampling, yielding the acquired measurement:
\begin{equation}
\mathbf{y} = \mathcal{P} \odot \mathbf{k}_{\mathrm{MB}} + \mathbf{n},
\label{eq:rw_measurement}
\end{equation}
where $\mathcal{P}$ is a Cartesian sampling mask, $\odot$ denotes element-wise multiplication, and $\mathbf{n}$ represents measurement noise.

SMS reconstruction aims to recover the slice-wise k-space
$\{\mathbf{k}_s\}_{s=1}^{B}$ from the collapsed measurement $\mathbf{y}$.
Due to the simultaneous presence of structured inter-slice aliasing and
k-space incompleteness, this inverse problem is highly ill-posed.
A generic regularized formulation can be written directly in k-space as
\begin{equation}
\label{eq:rw_generic_objective}
\begin{aligned}
\{\widehat{\mathbf{k}}_s\}_{s=1}^{B}
&=
\arg\min_{\{\mathbf{k}_s\}_{s=1}^{B}}
\;
\Bigl\lVert
\mathcal{P}\odot
\Bigl(\sum_{s=1}^{B}\mathcal{C}_s(\mathbf{k}_s)\Bigr)
-\mathbf{y}
\Bigr\rVert_2^2 \\
&\quad +\;
\lambda \,\mathcal{R}\!\left(\{\mathbf{k}_s\}_{s=1}^{B}\right),
\end{aligned}
\end{equation}
where $\mathcal{R}(\cdot)$ encodes prior information, such as sparsity-based compressed sensing~\cite{r33} and self-consistency-based parallel imaging methods such as SPIRiT~\cite{r10} and ESPIRiT~\cite{r11}, and $\lambda$ balances data fidelity and regularization.

This formulation highlights the intrinsic coupling between slice separation and in-plane completion, which becomes increasingly ill-conditioned under high multiband factors and strong in-plane acceleration.


\subsection{Learning-Based SMS Reconstruction}

The limitations of fixed linear operators under high acceleration have
motivated extensive research into learning-based SMS reconstruction
methods~\cite{r29,r27,r28,r30,r31}.
A representative calibration-guided learning approach is RAKI~\cite{r12}. 
RAKI trains a scan-specific convolutional network on ACS data and uses it to predict missing k-space samples from acquired ones.
Because the model is trained per scan, it does not require large external training datasets and can be integrated into practical SMS pipelines.
Representative SMS-specific learning and iterative frameworks include ROCK-SPIRiT~\cite{r13}, SMS-COOKIE~\cite{r14}, and multi-slice deep reconstruction models for dynamic/perfusion settings~\cite{r16,r17}.
Although effective under matched acquisition settings, these approaches do not explicitly parameterize or enforce SMS-specific deterministic physics, such as CAIPI-modulated slice superposition and structured inter-slice coupling, which can make their generalization sensitive to acquisition protocols and acceleration factors. 

More recently, diffusion-based generative models have been extended to MRI reconstruction and SMS reconstruction by enforcing the encoding model during sampling~\cite{r18,r19,r20,r21,r23,r24}.
Representative diffusion-based methods extend denoising diffusion models to SMS by enforcing the encoding model during sampling, for example through stepwise data-consistency projections~\cite{r25}. Some approaches operate in image space, while others formulate diffusion directly in k-space~\cite{r26}.
Despite these differences, most of them still construct the generative process by injecting stochastic noise and learning a denoiser to remove it. In contrast, SMS slice superposition and in-plane undersampling are deterministic degradations induced by known operators.
This mismatch creates a modeling conflict: SMS physics is often imposed as external data-consistency projections during inference, rather than being embedded in the generative dynamics. As a result, reconstructions under strong SMS and in-plane acceleration may still show structured slice leakage or artifacts.

This observation motivates diffusion formulations that model deterministic degradations directly.
In parallel, cold diffusion has been proposed to invert deterministic
transformations by constructing degradation chains without additive
noise~\cite{r22}.
Existing cold-diffusion-based MRI methods typically define the forward
degradation process through k-space masking or related acquisition
operators, while learning the reverse dynamics mainly in the image
domain.
This design is effective for modeling in-plane undersampling, but it
does not explicitly represent the deterministic slice superposition
inherent to SMS acquisitions.
As a result, the interaction between slice mixing and k-space
incompleteness is not directly modeled in the generative process,
limiting the ability of existing cold diffusion formulations to address
structured slice leakage under high SMS acceleration.

\section{Method}
\label{sec:method}

\subsection{Motivation}
\label{sec:method_overview}

In SMS MRI, both CAIPI-modulated slice superposition and Cartesian in-plane undersampling are deterministic effects governed by known operators.
This motivates an operator-aligned inference trajectory that explains each intermediate state via structured interference and missing-data corruption, rather than assuming additive Gaussian noise.
A practical challenge in SMS reconstruction is that two deterministic degradations are
strongly coupled: Structured inter-slice interference from SMS superposition and
missing k-space data from in-plane undersampling.
Directly solving both effects in a single inference stage is often ill-conditioned and
can lead to unstable trade-offs between slice leakage suppression and high-frequency
recovery.
To reduce this coupling, we decompose the reconstruction into two chained stages.
Stage-M focuses on resolving CAIPI-induced inter-slice interference, while Stage-U
addresses in-plane completion under explicit frequency-domain constraints.

\begin{figure*}[t]
  \centering
  \includegraphics[width=0.90\textwidth]{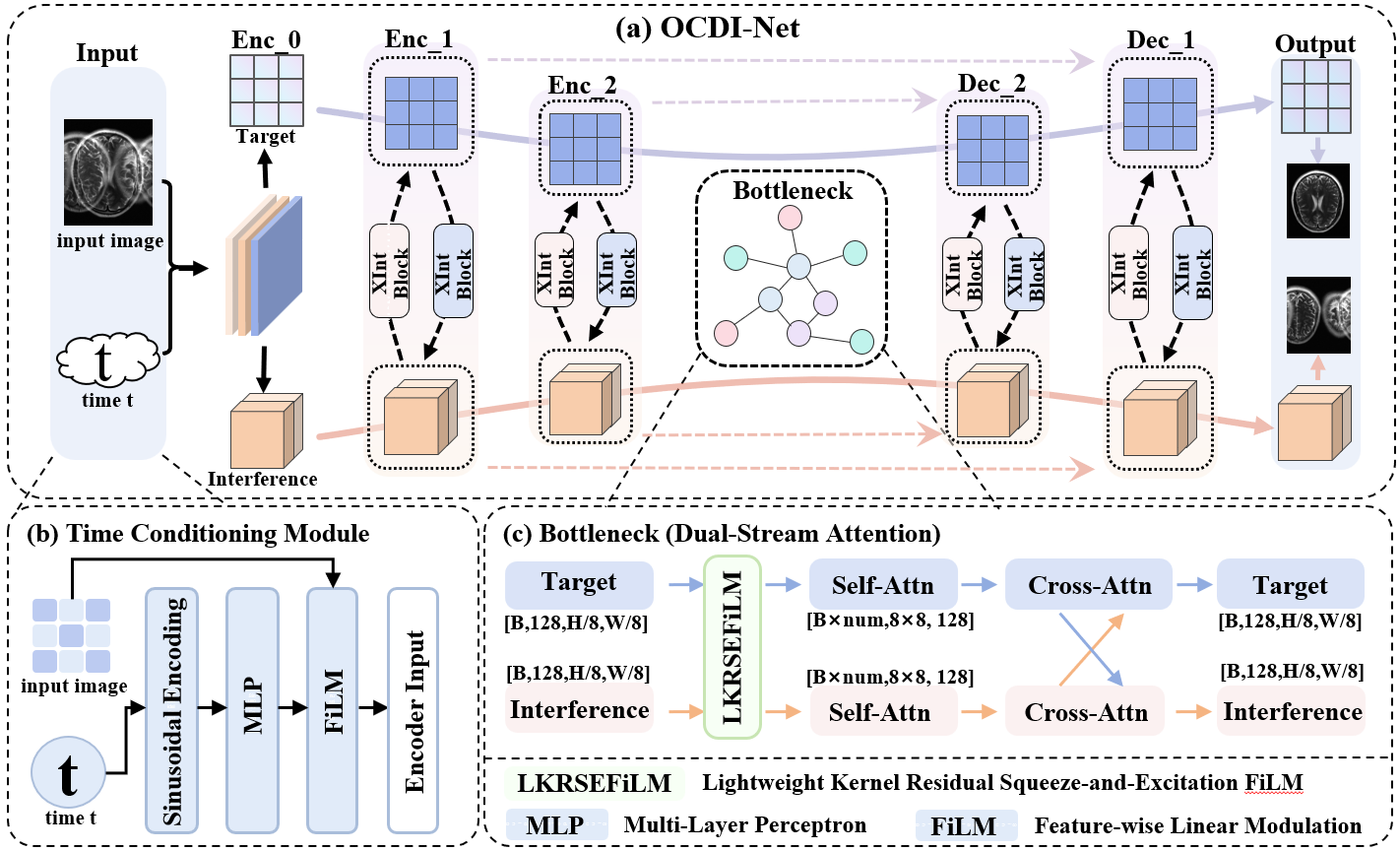}
  \caption{Architecture of the operator-conditional dual-stream interaction network (OCDI-Net).
  OCDI-Net maintains a \emph{target-content stream} and an \emph{interference stream} throughout a U-Net-style encoder--decoder~\cite{r15}.
  Cross-stream interaction blocks exchange information under conditioning on diffusion step $t$ and operator/stage indicator $\Omega$.
  The network outputs the predicted degradation $\widehat{\mathbf{d}}_t$ used by the operator-guided reverse updates in Sec.~\ref{sec:method_cold_formulation}.}
  \label{fig:ocdi}
\end{figure*}

\subsection{Operator-Guided Deterministic Trajectory}
\label{sec:method_cold_formulation}

Let $\mathbf{k}^{\star}$ denote the clean k-space variable of interest at the current stage.
We use a stage indicator $\Omega \in \{M,U\}$ to specify the operator setting, where
$\Omega=M$ corresponds to SMS slice separation and $\Omega=U$ corresponds to in-plane completion.
Our deterministic trajectory is constructed between $\mathbf{k}^{\star}$ and a physically consistent terminal state
$\mathbf{y}_{\Omega}$, which we write in a unified form as
\begin{equation}
\mathbf{y}_{\Omega} \triangleq \mathcal{A}_{\Omega}(\mathbf{k}^{\star}).
\end{equation}

For Stage-U, $\mathcal{A}_{\Omega}$ reduces to the Cartesian undersampling operator,
$\mathcal{A}_{U}(\mathbf{k}^{\star}) \triangleq \mathcal{P}\odot \mathbf{k}^{\star}$.
For Stage-M, the terminal state is a CAIPI-aligned collapsed k-space for a target slice $s^\star$,
expressed in the target-aligned coordinate system as
\begin{equation}
\mathcal{A}_{M}(\mathbf{k}^{\star})
\triangleq
\widetilde{\mathbf{y}}_{s^\star}
=
\mathbf{k}_{s^\star}
+
\sum_{s\neq s^\star}
\bigl(\mathcal{C}_{s^\star}^{-1}\mathcal{C}_{s}\bigr)
(\mathbf{k}_{s}),
\end{equation}
where $\mathcal{C}_{s}(\cdot)$ denotes the known CAIPI modulation operator and
$\mathbf{k}^{\star}\equiv \mathbf{k}_{s^\star}$.
Note that $\widetilde{\mathbf{y}}_{s^\star}$ is not generated by applying a known operator to the target slice alone;
instead, it contains an interference realization contributed by other simultaneously excited slices, whose structure
is governed by the known relative CAIPI operators
$\{\mathcal{C}_{s^\star}^{-1}\mathcal{C}_{s}\}_{s\neq s^\star}$.

With $\mathbf{y}_{\Omega}$ defined above, we introduce the corresponding structured degradation term:
\begin{equation}
\mathbf{d}_{\Omega}
\triangleq
\mathbf{y}_{\Omega}-\mathbf{k}^{\star}
=
\mathcal{A}_{\Omega}(\mathbf{k}^{\star})-\mathbf{k}^{\star}.
\end{equation}

Accordingly, $\mathbf{d}_{M}$ represents operator-governed inter-slice interference in the target-aligned space,
while $\mathbf{d}_{U}$ represents mask-induced missing-data corruption.

\subsubsection{Deterministic Forward Process}
\label{sec:method_forward_new}

We define diffusion steps $t \in \{0,1,\cdots,T\}$ with a monotone schedule $\alpha_t \in [0,1]$ satisfying $\alpha_0 = 0$ and $\alpha_T = 1$.
The forward state is defined as a deterministic interpolation between the clean target $\mathbf{k}^{\star}$
and the stage-consistent terminal state $\mathbf{y}_{\Omega}$:
\begin{equation}
\mathbf{x}_t
=
(1-\alpha_t)\,\mathbf{k}^{\star}
+
\alpha_t\,\mathcal{A}_{\Omega}(\mathbf{k}^{\star})
=
\mathbf{k}^{\star}
+
\alpha_t\,\mathbf{d}_{\Omega},
\label{eq:cold_forward_interp}
\end{equation}
under this construction, $\mathbf{x}_0=\mathbf{k}^{\star}$ and $\mathbf{x}_T=\mathbf{y}_{\Omega}$, meaning the terminal state corresponds to a physically meaningful corrupted state for the given stage, rather than a Gaussian-noise-corrupted sample.

\subsubsection{Reverse Update via Degradation Prediction}
\label{sec:method_reverse_new}

Given an intermediate state $\mathbf{x}_t$, a neural network $\mathcal{F}_{\theta}$ predicts the structured degradation
$\widehat{\mathbf{d}}_t$ conditioned on the diffusion step and operator/stage indicator:
\begin{equation}
\widehat{\mathbf{d}}_t
=
\mathcal{F}_{\theta}(\mathbf{x}_t, t;\Omega).
\label{eq:pred_degradation}
\end{equation}

Consistent with Eq.~\eqref{eq:cold_forward_interp}, $\widehat{\mathbf{d}}_t$ aims to explain the current state in terms of the stage-specific structured degradation $\mathbf{d}_{\Omega}$.
From Eq.~\eqref{eq:cold_forward_interp}, we have $\mathbf{x}_t=\mathbf{k}^{\star}+\alpha_t\mathbf{d}_{\Omega}$, which implies the identity $\mathbf{k}^{\star}=\mathbf{x}_t-\alpha_t\mathbf{d}_{\Omega}$.
Therefore, if $\widehat{\mathbf{d}}_t \approx \mathbf{d}_{\Omega}$, a natural estimate of the clean target at step $t$ is obtained by subtracting the predicted degradation scaled by $\alpha_t$:
\begin{equation}
\widehat{\mathbf{k}}^{(t)}
=
\mathbf{x}_t
-
\alpha_t\,\widehat{\mathbf{d}}_t,
\label{eq:clean_estimate}
\end{equation}
and the next state follows the same forward form $\mathbf{x}_{t-1}=\mathbf{k}^{\star}+\alpha_{t-1}\mathbf{d}_{\Omega}$ by reusing the same degradation estimate:
\begin{equation}
\mathbf{x}_{t-1}
=
\widehat{\mathbf{k}}^{(t)}
+
\alpha_{t-1}\,\widehat{\mathbf{d}}_t,
\label{eq:reverse_propagate}
\end{equation}
where $t$ indexes the current reverse iteration step and decreases from $T$ to $1$, $t-1$ denotes the next step, and $\{\alpha_t\}$ is the predefined monotone schedule.

This deterministic update can be interpreted as an operator-aligned analogue of implicit diffusion sampling, where the network progressively explains $\mathbf{x}_t$ using physically grounded structured degradations, rather than denoising Gaussian perturbations.
The stage indicator $\Omega$ enables the same formulation to handle both operator-governed inter-slice interference removal and mask-induced missing-data recovery within a unified framework.

\subsection{OCDI-Net as Degradation Predictor}
\label{sec:method_ocdi_pred}

In the proposed operator-guided framework, the neural network is not a generic denoiser.
Instead, it predicts the operator-induced structured degradation term that arises from SMS superposition and in-plane undersampling.
This design tightly aligns network prediction with the physically grounded degradation term in Sec.~\ref{sec:method_cold_formulation} and enables principled inversion under strong acceleration.

\subsubsection{Predicting Operator-induced Degradation}
\label{sec:method_ocdi_predict}
As derived in Sec.~\ref{sec:method_cold_formulation}, the deterministic inversion trajectory yields
$\mathbf{x}_t = \mathbf{k}^{\star} + \alpha_t \mathbf{d}_{\Omega}$,
where $\mathbf{d}_{\Omega} \triangleq \mathcal{A}_{\Omega}(\mathbf{k}^{\star}) - \mathbf{k}^{\star}$
denotes the operator-induced structured degradation associated with the acquisition operator
$\mathcal{A}_{\Omega}$.
Accordingly, OCDI-Net predicts the degradation term using Eq.~\eqref{eq:pred_degradation},
rather than estimating a stochastic noise sample or a score function.
This formulation is suited to SMS MRI, in which aliasing and slice leakage arise as
deterministic interference patterns governed by CAIPI encoding and sampling masks, rather than as
random perturbations.

\subsubsection{Network Architecture and Operator-Consistent Design}
\label{sec:method_ocdi_arch_overview}
A standard neural network processes the input k-space as a single mixed signal. However, this input contains two distinct components: the consistent anatomy of the target slice and the structured aliasing from other slices. If a network processes them together in one stream, it often struggles to distinguish true anatomical details from leakage artifacts. To address this, we design a dual-stream architecture. We explicitly force the network to separate the features into two independent paths: one for the target content and one for the interference patterns.

Fig.~\ref{fig:ocdi} illustrates the overall architecture of OCDI-Net.
The network adopts a U-Net-style encoder--decoder backbone that operates directly on complex-valued k-space states $\mathbf{x}_t$.
In practice, $\mathbf{x}_t$ is represented using real and imaginary channels, and is processed at multiple spatial resolutions.
This multi-scale design enables the network to simultaneously capture local interpolation structure and global aliasing patterns induced by the acquisition operators.
A time-conditioning pathway injects the diffusion step $t$ into the entire hierarchy, allowing the same architecture to act as a step-dependent predictor of degradation along the deterministic inversion  trajectory.

Beyond time conditioning, OCDI-Net is explicitly designed to be aware of the underlying acquisition operator.
As shown in Fig.~\ref{fig:ocdi}(b), the step index $t$ is embedded using a sinusoidal positional encoding followed by an MLP and converted into feature-wise modulation parameters, which are applied throughout the encoder and decoder blocks.
In addition, we provide a stage indicator $\Omega \in \{M, U\}$ to differentiate between Stage-M and Stage-U.
This allows a single backbone to specialize its behavior to different deterministic degradation mechanisms, including CAIPI-modulated inter-slice interference and Cartesian undersampling.
As a result, the prediction is used inside the reverse updates in Eq.~\eqref{eq:clean_estimate}--\eqref{eq:reverse_propagate},
instead of learning a direct mapping from $\mathbf{x}_t$ to $\mathbf{k}^{\star}$ that ignores the operator setting.

A central challenge in SMS reconstruction is that the observed k-space state contains two qualitatively different components:
Slice-consistent target content and structured inter-slice interference whose structure is dictated by CAIPI phase modulation and sampling geometry.
If these components are represented within a single latent space, slice-consistent structures and leakage patterns can become entangled, often leading to residual slice leakage under high multiband factors and strong in-plane acceleration.
To reflect this physical composition, OCDI-Net maintains two asymmetric feature streams at every resolution level, as illustrated in Fig.~\ref{fig:ocdi}(a).
Specifically, a target-content stream $\mathbf{z}^{\mathrm{T}}$ focuses on slice-consistent information, while an interference stream $\mathbf{z}^{\mathrm{I}}$ is dedicated to structured aliasing and leakage components.
After each encoder stage produces multi-scale features, they are split into $(\mathbf{z}^{\mathrm{T}}, \mathbf{z}^{\mathrm{I}})$ (e.g., by channel partition), propagated in parallel through the multi-resolution U-Net pathway, and preserved in the decoder.
This dual-stream organization aligns the network’s internal representation with the operator-induced decomposition in $\mathbf{d}_\Omega$, encouraging $\mathcal{F}_\theta$ to explain $\mathbf{x}_t$ using deterministic interference patterns rather than absorbing them into target content.

Although the two streams specialize in different components, accurate reconstruction requires controlled information exchange between them.
The target stream must incorporate contextual cues about leakage structure to effectively suppress interference, while the interference stream must remain informed by anatomical support to avoid generating inconsistent artifacts.
Accordingly, OCDI-Net introduces cross-stream interaction blocks at multiple scales, conditioned on both the diffusion step and the operator type:
\begin{equation}
(\mathbf{z}^{\mathrm{T}}_{\ell}, \mathbf{z}^{\mathrm{I}}_{\ell})
\leftarrow
\Phi_{\ell}\!\left(
\mathbf{z}^{\mathrm{T}}_{\ell},
\mathbf{z}^{\mathrm{I}}_{\ell};
\mathbf{e}_t,\Omega
\right),
\label{eq:ocdi_interaction}
\end{equation}
where $\mathbf{e}_t$ denotes the time-step embedding.
In practice, $\Phi_\ell$ first refines each stream using convolutional blocks and self-attention blocks~\cite{r35}.
It then performs cross-stream interaction using an attention-based gating module, which exchanges complementary information between $\mathbf{z}^{\mathrm{T}}$ and $\mathbf{z}^{\mathrm{I}}$.
At the bottleneck (Fig.~\ref{fig:ocdi}(c)), a dual-stream attention module further captures global dependencies and explicit cross-stream coupling, which is particularly important for SMS because CAIPI-induced interference can be spatially structured and globally coherent across k-space.
Conditioning these interactions on $(t,\Omega)$ allows a single architecture to adapt its interaction patterns across both stages, emphasizing inter-slice disentanglement in Stage-M and high-frequency completion under strict frequency-domain constraints in Stage-U.

Finally, the decoder aggregates multi-scale information within each stream and outputs the predicted degradation $\widehat{\mathbf{d}}_t$ directly in k-space (Fig.~\ref{fig:ocdi}(a), output head).
This predicted degradation is then inserted into the operator-guided reverse update rule in Eq.~\eqref{eq:clean_estimate}--\eqref{eq:reverse_propagate}.
Consequently, the learning component focuses exclusively on explaining and removing operator-induced degradations, while the overall reconstruction algorithm enforces strict physical consistency through the known acquisition operators.

\subsubsection{Training Objective}
\label{sec:method_ocdi_train}

We train OCDI-Net by sampling a diffusion step $t$ and constructing $\mathbf{x}_t$ using Eq.~\eqref{eq:cold_forward_interp}.
The primary supervision is applied in k-space using an $\ell_1$ loss:
\begin{equation}
\mathcal{L}_k = \left\| \widehat{\mathbf{k}}^{(t)} - \mathbf{k}^{\star} \right\|_{1},
\label{eq:loss_k_new}
\end{equation}
optionally augmented with an image-domain loss on coil-combined magnitude:
\begin{equation}
\mathcal{L}_I
=
\left\|
\; \big| \mathcal{B}(\widehat{\mathbf{k}}^{(t)}) \big|
-
\big| \mathcal{B}(\mathbf{k}^{\star}) \big|
\;\right\|_{1},
\label{eq:loss_I_new}
\end{equation}
where $\mathcal{B}(\cdot)$ denotes inverse Fourier transform followed by coil combination.
The total loss is
\begin{equation}
\mathcal{L} = \lambda_k \mathcal{L}_k + \lambda_I \mathcal{L}_I.
\label{eq:loss_total_new}
\end{equation}

\subsection{Two-Stage Inference }
\label{sec:method_twostage}

\begin{figure}[t]
  \centering
  \includegraphics[width=0.5\textwidth]{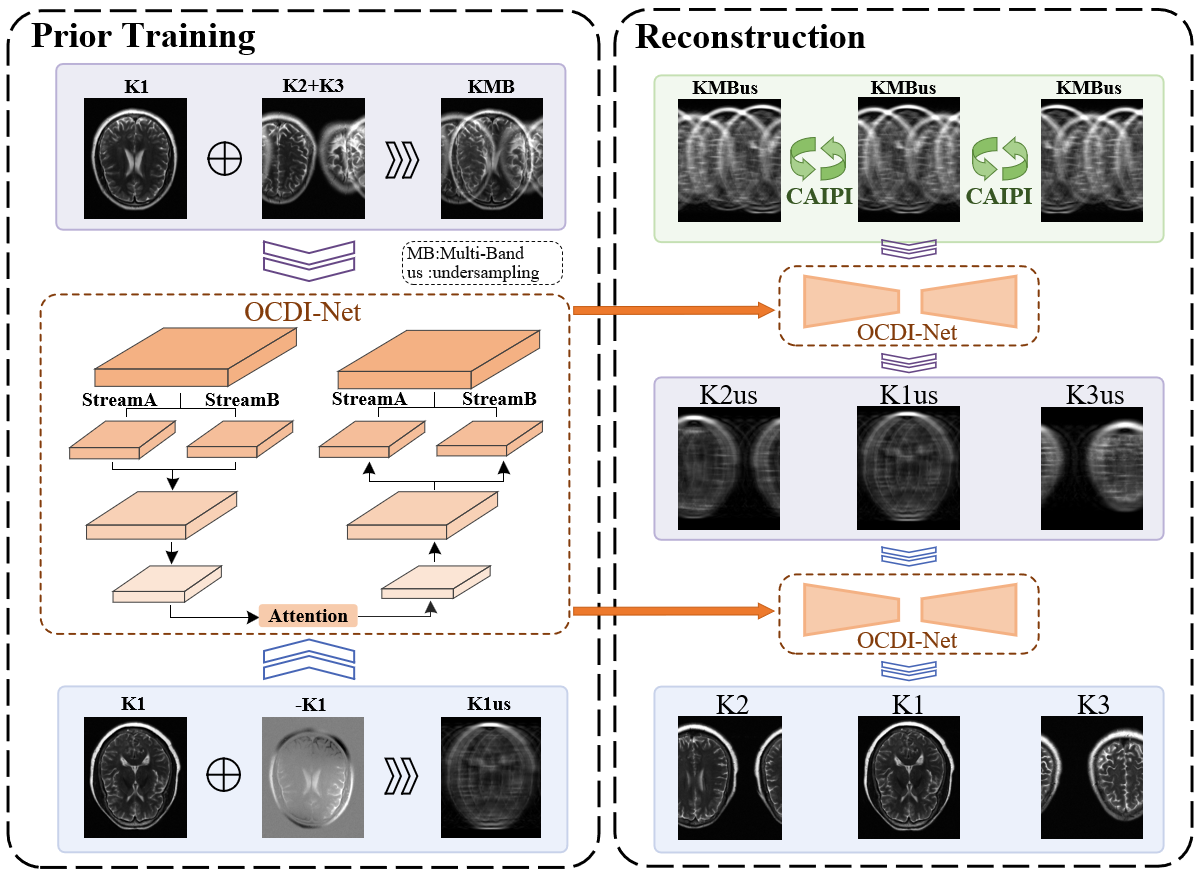}
  \caption{Overall pipeline of the proposed operator-guided SMS reconstruction framework.
$K_1$, $K_2$, and $K_3$ denote individual slice k-space components, and $K_{\mathrm{MB}}$ denotes their CAIPI-modulated multiband superposition.
The suffix ``us'' indicates in-plane undersampled k-space.
During reconstruction, each target slice is aligned by inverse CAIPI modulation and reconstructed by two chained stages using the same OCDI-Net: Stage-M performs slice separation, followed by Stage-U for in-plane completion.}
  \label{fig:pipeline}
\end{figure}

As illustrated in Fig.~\ref{fig:pipeline}, we instantiate the proposed operator-guided deterministic inversion formulation into two chained stages to decouple structured inter-slice disentanglement from missing-data recovery.
This design follows the physical origins of degradations in SMS MRI: CAIPI-modulated slice superposition produces structured interference, while in-plane acceleration induces deterministic k-space incompleteness.
The two stages share the same operator-guided deterministic reverse update,
but they use different stage-specific degradation terms and different consistency constraints during sampling.

In real acquisitions, the measured data consist of the collapsed SMS k-space $\mathbf{y}$ and the sampling mask $\mathcal{P}$.
The CAIPI modulation operators $\{\mathcal{C}_s\}_{s=1}^{B}$ are known from the sequence prescription.
Slice-wise fully sampled k-space $\{\mathbf{k}_s\}_{s=1}^{B}$ and the corresponding images $\{\mathbf{x}_s\}_{s=1}^{B}$ are unavailable at inference; they are used only to synthesize training pairs in retrospective experiments.
To stabilize inference under strong acceleration, we estimate a slice-wise low-frequency k-space anchor using a linear kernel-based SMS reconstruction.
This anchor is extracted from the central k-space region and is used only during inference to guide low-frequency components.

\subsubsection{Stage-M}
Stage-M targets structured inter-slice interference removal under SMS superposition.
A key distinction in this stage is between the unmasked collapsed multiband k-space and the acquired measurement.
We denote the unmasked collapsed multiband k-space as $\widetilde{\mathbf{k}}_{\mathrm{MB}}$, which follows Eq.~\eqref{eq:rw_sms_superposition}.
When in-plane undersampling is applied, the acquired measurement satisfies
$\mathbf{y} = \mathcal{P}\odot \widetilde{\mathbf{k}}_{\mathrm{MB}}$.

Accordingly, the initialization of Stage-M depends on the sampling mask. 
When $\mathcal{P} = \mathbf{1}$, which denotes an all-one mask, Stage-M starts from the unmasked collapsed k-space, $\mathbf{x}^{(M)}_{T}=\widetilde{\mathbf{k}}_{\mathrm{MB}}$.
When $\mathcal{P}\neq\mathbf{1}$, Stage-M starts from the acquired measurement,
$\mathbf{x}^{(M)}_{T}=\mathbf{y}$,
where the unacquired entries are zeros.
As a result, the output of Stage-M may still be in-plane undersampled when $\mathcal{P}\neq\mathbf{1}$, and a subsequent Stage-U is required for in-plane completion.

For clarity, we illustrate the formulation using the MB$=3$ case.
The collapsed multiband k-space can be rewritten as
\begin{equation}
\mathbf{k}_{\mathrm{MB}} = \mathbf{k}_{1} + \mathcal{C}_{2}(\mathbf{k}_{2}) + \mathcal{C}_{3}(\mathbf{k}_{3}),
\end{equation}
where $\mathbf{k}_{1}$ denotes the target slice and $\mathbf{k}_{2}, \mathbf{k}_{3}$ denote the remaining simultaneously excited slices.
The superposed non-target contribution is treated as a structured interference term:
\begin{equation}
\mathbf{d}_{\mathrm{M}} \triangleq \mathcal{C}_{2}(\mathbf{k}_{2}) + \mathcal{C}_{3}(\mathbf{k}_{3}),
\end{equation}
which is deterministic given the simultaneously excited slices and CAIPI encoding.
Stage-M therefore applies the operator-guided reverse updates in Sec.~\ref{sec:method_cold_formulation} with
$\mathbf{d}=\mathbf{d}_{\mathrm{M}}$ to estimate the target-slice k-space $\widehat{\mathbf{k}}_{1}$. The same applies when the target is $\mathbf{k}_{2}$ and $\mathbf{k}_{3}$.

\begin{algorithm}[t]
\caption{Two-Stage Operator-Guided deterministic inversion Inference (Stage-M $\rightarrow$ Stage-U)}
\label{alg:two_stage_infer}
\footnotesize
\begin{algorithmic}[1]
\Require Collapsed measurement $\mathbf{y}$; in-plane mask $\mathcal{P}$; CAIPI operators $\{\mathcal{C}_s\}$;
diffusion schedule $\{\alpha_t\}$; trained predictor $\mathcal{F}_{\theta}(\mathbf{x}, t;\Omega)$.
Optional: low-frequency k-space anchor
$\mathbf{k}^{\mathrm{LF}}_{s^\star}$ estimated by a linear kernel-based method;
guidance interval $G$.
\Ensure Stage-M estimate $\widehat{\mathbf{k}}_{s^\star}$ and final reconstruction $\widehat{\mathbf{k}}^{\mathrm{full}}_{s^\star}$.

\State \textbf{Stage-M (slice separation)}
\State Initialize $\mathbf{x}^{(M)}_{T_M} \gets \widetilde{\mathbf{y}}_{s^\star}$
\Comment{CAIPI-aligned collapsed k-space}
\For{$t=T_M$ \textbf{down to} $1$}
    \State $\widehat{\mathbf{d}}_t \gets \mathcal{F}_{\theta}(\mathbf{x}^{(M)}_{t}, t; M)$
    \State $\widehat{\mathbf{k}}^{(t)} \gets \mathbf{x}^{(M)}_{t} - \alpha_t \widehat{\mathbf{d}}_t$
    \State $\mathbf{x}^{(M)}_{t-1} \gets \widehat{\mathbf{k}}^{(t)} + \alpha_{t-1}\widehat{\mathbf{d}}_t$
\EndFor
\State $\widehat{\mathbf{k}}_{s^\star} \gets \mathbf{x}^{(M)}_{0}$
\vspace{2pt}
\State \textbf{Stage-U (in-plane completion)}
\State $\mathbf{x}^{(U)}_{T_U} \gets \widehat{\mathbf{k}}_{s^\star}$ \Comment{warm start from Stage-M}
\For{$t=T_U$ \textbf{down to} $1$}
    \State $\widehat{\mathbf{d}}_t \gets \mathcal{F}_{\theta}(\mathbf{x}^{(U)}_{t}, t; U)$
    \State $\widehat{\mathbf{k}}^{(t)} \gets \mathbf{x}^{(U)}_{t} - \alpha_t \widehat{\mathbf{d}}_t$
    \State $\mathbf{x}^{(U)}_{t-1} \gets \widehat{\mathbf{k}}^{(t)} + \alpha_{t-1}\widehat{\mathbf{d}}_t$
    
    \State $\mathbf{x}^{(U)}_{t-1} \gets \mathcal{P}\odot \widehat{\mathbf{k}}_{s^\star} + (1-\mathcal{P})\odot \mathbf{x}^{(U)}_{t-1}$  
    \Comment{DC w.r.t. pseudo-measurement from Stage-M}

    \If{Low-frequency k-space anchor is available \textbf{and} $(t \bmod G = 0)$}
        \State $\mathbf{x}^{(U)}_{t-1} \gets \mathcal{M}_{\mathrm{ACS}}\odot \mathbf{k}^{\mathrm{LF}}_{s^\star} + (1-\mathcal{M}_{\mathrm{ACS}})\odot \mathbf{x}^{(U)}_{t-1}$
    \EndIf
\EndFor
\State $\widehat{\mathbf{k}}^{\mathrm{full}}_{s^\star} \gets \mathbf{x}^{(U)}_{0}$
\State \Return $\widehat{\mathbf{k}}_{s^\star}$, $\widehat{\mathbf{k}}^{\mathrm{full}}_{s^\star}$
\end{algorithmic}
\end{algorithm}

\subsubsection{Stage-U}
In Stage-U, we treat the in-plane undersampling operator as the stage-specific degradation and apply the same operator-guided reverse updates described in Sec.~\ref{sec:method_cold_formulation}.
In real SMS acquisitions, slice-wise fully sampled k-space $\mathbf{y}_{s^\star}$ is unavailable, which precludes direct data-consistency enforcement at the slice level.
To address this, we use the Stage-M output as a pseudo-measurement on the acquired in-plane locations, defined as
\begin{equation}
\widetilde{\mathbf{y}}_{s^\star} \triangleq \mathcal{P}\odot \widehat{\mathbf{k}}_{s^\star}.
\label{eq:pseudo_meas}
\end{equation}

This pseudo-measurement enables data-consistency projection  while avoiding undefined slice-wise measurements.

In practice, we adopt a conditioned warm start by initializing Stage-U with the Stage-M output,
$\mathbf{x}^{(U)}_{T} \triangleq \widehat{\mathbf{k}}_{s^\star}$,
and iteratively refine it to recover missing high-frequency components.
After each reverse update, data consistency is enforced on the acquired locations using the pseudo-measurement:
\begin{equation}
\mathbf{x}_{t-1}
\leftarrow
\mathcal{P}\odot \widehat{\mathbf{k}}_{s^\star}
+
(1-\mathcal{P})\odot \mathbf{x}_{t-1}.
\end{equation}

\subsubsection{Chained Inference}
At inference time, we first run Stage-M to obtain $\widehat{\mathbf{k}}_{s^\star}$ and then run Stage-U to obtain the final reconstruction, as summarized in Fig.~\ref{fig:pipeline}.
The complete two-stage inference procedure is summarized in Algorithm~\ref{alg:two_stage_infer}.

\FloatBarrier



\section{Experiments}
\label{sec:experiments}
\subsection{Experimental Setup}
\label{sec:exp_setup}
\subsubsection{Datasets}
\label{sec:exp_datasets}

The proposed method evaluate the proposed method on both retrospective and prospective SMS datasets.
For retrospective experiments, we use the public fastMRI brain dataset~\cite{r32}, which provides fully sampled k-space measurements for T2-weighted brain scans. All fastMRI experiments use the 16-coil brain dataset.
The dataset is split at the scan level into mutually exclusive training, validation, and testing subsets with an approximate ratio of 8:1:1.
We extract 12 representative axial slices per volume.
Retrospective SMS data are generated directly in k-space by applying slice-dependent CAIPI phase modulation followed by linear slice summation.
For a multiband factor of three, standard CAIPI shifts of $0$ and $\pm \mathrm{FOV}/3$ along the phase-encoding direction are used.
In-plane acceleration is simulated using uniform Cartesian undersampling with $R \in \{1,2,3\}$, and 32 central phase-encoding lines are retained as ACS for all settings.
The same undersampling masks and ACS configuration are applied across all compared methods.

For prospective validation, we further evaluate the proposed method on a private DWI dataset acquired at Tianjin University (TJU).
All scans are acquired on a uMR~15NX Frontier 3.0~T MRI scanner using single-shot echo-planar imaging, with $1.25~\mathrm{mm}$ isotropic resolution, matrix size $360 \times 180 \times 36$, and both anterior--posterior and posterior--anterior phase-encoding directions.

For training on DWI, only single-band acquisitions with in-plane acceleration factor $R=2$ at $b=1000~\mathrm{s/mm^2}$ are used.
These data are reconstructed using conventional GRAPPA and serve as surrogate slice-wise references for learning-based SMS reconstruction.
For testing, prospectively acquired SMS data with $\mathrm{MB}=2$ and $R=2$ at $b=1000~\mathrm{s/mm^2}$ are used, where fully sampled slice-wise references are unavailable.
To assess cross-contrast generalization, the same trained model is additionally evaluated on $\mathrm{MB}=2$, $R=2$ data acquired at $b=0~\mathrm{s/mm^2}$ from the same scanner, without any retraining. Our implementation has been publicly released at: https://github.com/yqx7150/OCDI.

\subsubsection{Compared Methods}
\label{sec:exp_baselines}

We compare the proposed method with representative conventional and
learning-based approaches for SMS reconstruction.
The selected baselines cover image-domain, k-space-based, calibration-guided,
and diffusion-based reconstruction paradigms.
Specifically, SENSE is included as a classical image-domain parallel imaging
method that performs SMS unaliasing using coil sensitivity maps estimated from
central calibration data.
Slice-GRAPPA and RAKI represent k-space-based reconstruction approaches, where
Slice-GRAPPA relies on linear kernel interpolation calibrated from
low-frequency data, while RAKI replaces linear interpolation with a
scan-specific neural network trained directly on calibration data.
In addition, ROGER is included as a representative diffusion-based SMS reconstruction method formulated under an additive Gaussian noise model, which incorporates SMS encoding through SENSE-style forward operators and enforces
data consistency during the reverse diffusion process. We follow the official implementation and recommended hyperparameters, and apply the same SMS encoding operators and sampling masks as in other baselines.
All compared methods use identical SMS folding rules, undersampling masks,
and in-plane sampling configurations to ensure a fair comparison.

\subsubsection{Implementation Details and Evaluation Metrics}
\label{sec:exp_impl}

The proposed method is trained using retrospectively simulated SMS data, with separate training settings for retrospective fastMRI experiments and prospective TJU DWI experiments.
For fastMRI evaluations, training data are constructed from the public fastMRI brain dataset by applying CAIPI-modulated slice superposition and Cartesian in-plane undersampling in k-space.
For prospective DWI evaluations, the network is trained using single-band DWI data at $b=1000~\mathrm{s/mm^2}$, from which retrospective SMS samples are generated using the same CAIPI modulation and MB2R2 sampling pattern as in testing.
The trained model is then applied to prospectively acquired DWI MB2R2 data at $b=1000~\mathrm{s/mm^2}$, and the same model is directly tested on DWI MB2R2 data at $b=0~\mathrm{s/mm^2}$ to evaluate cross-contrast generalization, without any retraining.
Quantitative evaluation is conducted using peak signal-to-noise ratio (PSNR), structural similarity index measure (SSIM)~\cite{r38}, and normalized mean squared error (NMSE), computed on root-sum-of-squares magnitude images after inverse Fourier transformation and coil combination.
A consistent intensity normalization strategy is applied across all methods prior to metric computation to ensure fair comparison.
For prospective experiments without fully sampled references, qualitative comparisons are reported.

\subsection{Comparison with State-of-the-Arts}
\label{sec:sota}

\subsubsection{Visual Comparisons on fastMRI}
\label{sec:results_visual}

\begin{figure*}[!t]
  \centering
  \includegraphics[width=0.85\textwidth]{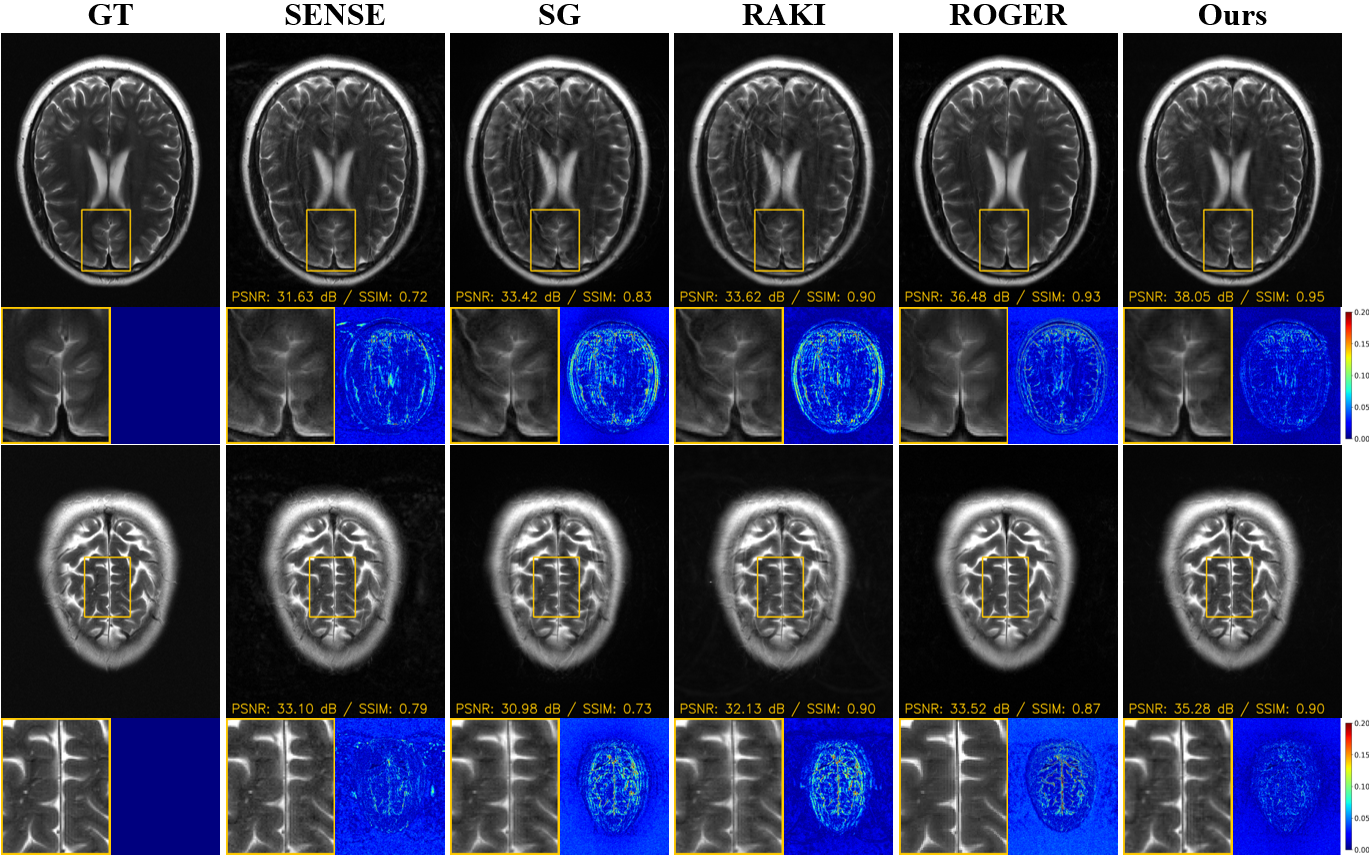}
  \vspace{-2mm}
  \caption{Visual comparisons on fastMRI under MB3R2.
Columns show GT and reconstructions by SENSE, Slice-GRAPPA (SG), RAKI, ROGER, and the proposed method.
Each example shows the full-FOV RSS image with ROI and the zoomed ROI with absolute error map.}

  \label{fig:fastmri_mb3r2}
  \vspace{-3mm}
\end{figure*}

Fig.~\ref{fig:fastmri_mb3r2} shows representative visual comparisons under the MB3R2
setting.
SENSE exhibits pronounced noise amplification and residual structured artifacts,
which blur cortical boundaries and suppress fine anatomical details.
Slice-GRAPPA alleviates part of the aliasing but still shows residual slice leakage and
loss of high-frequency information.
RAKI improves visual sharpness compared with linear baselines but introduces
non-uniform residual artifacts in challenging regions.
ROGER produces visually smoother reconstructions; however, structured errors remain around complex anatomical boundaries.This behavior is consistent with its Gaussian-noise diffusion formulation, which tends to trade local texture for smoothness when the dominant corruption is operator-governed and structured rather than random.
In contrast, OCDI produces reconstructions that are visually closest to the ground truth, with substantially reduced structured errors.
This improvement suggests more effective suppression of coherent inter-slice interference while preserving high-frequency  slice-consistent details.

\begin{figure*}[!t]
  \centering
  \includegraphics[width=0.45\textwidth]{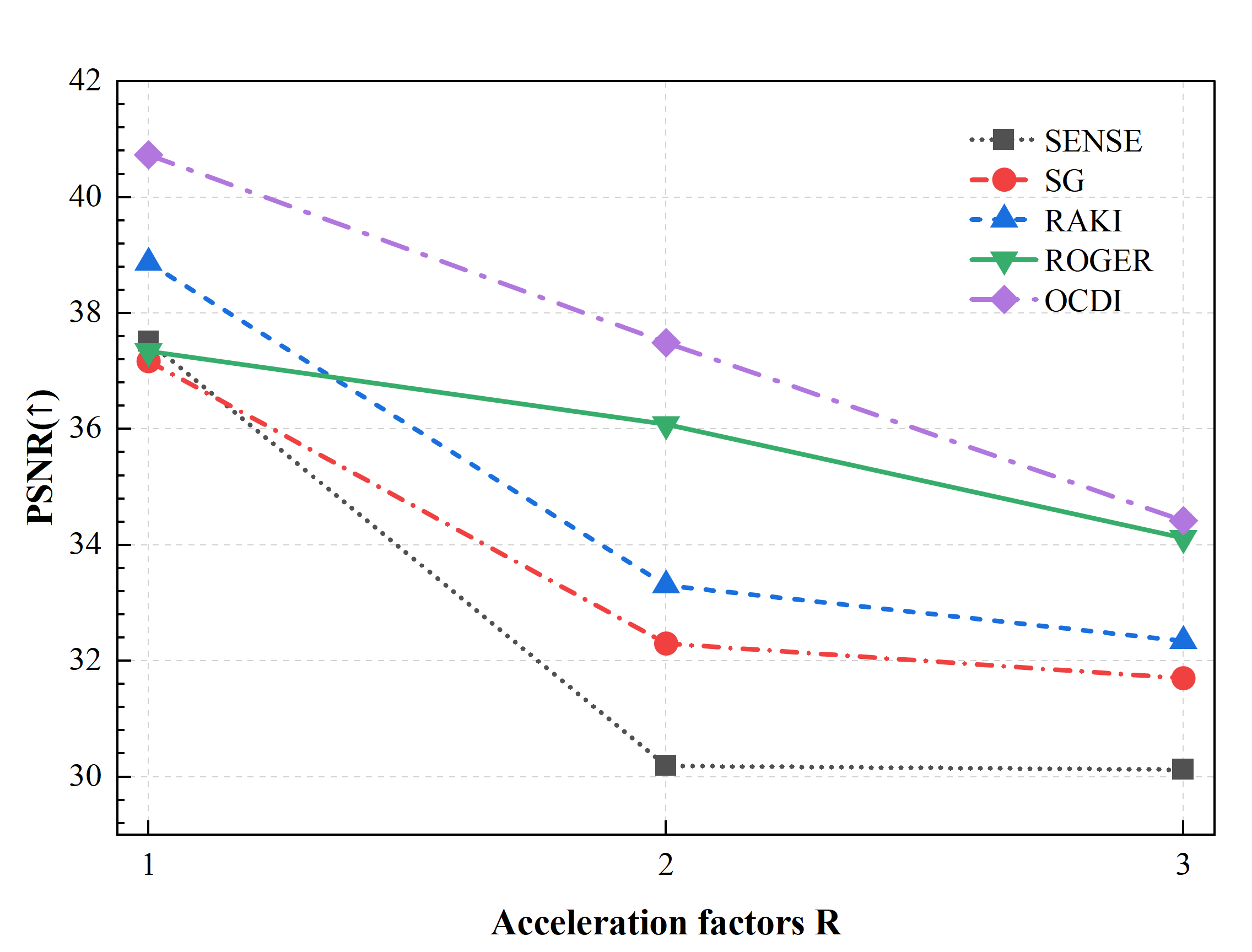}
  \includegraphics[width=0.45\textwidth]{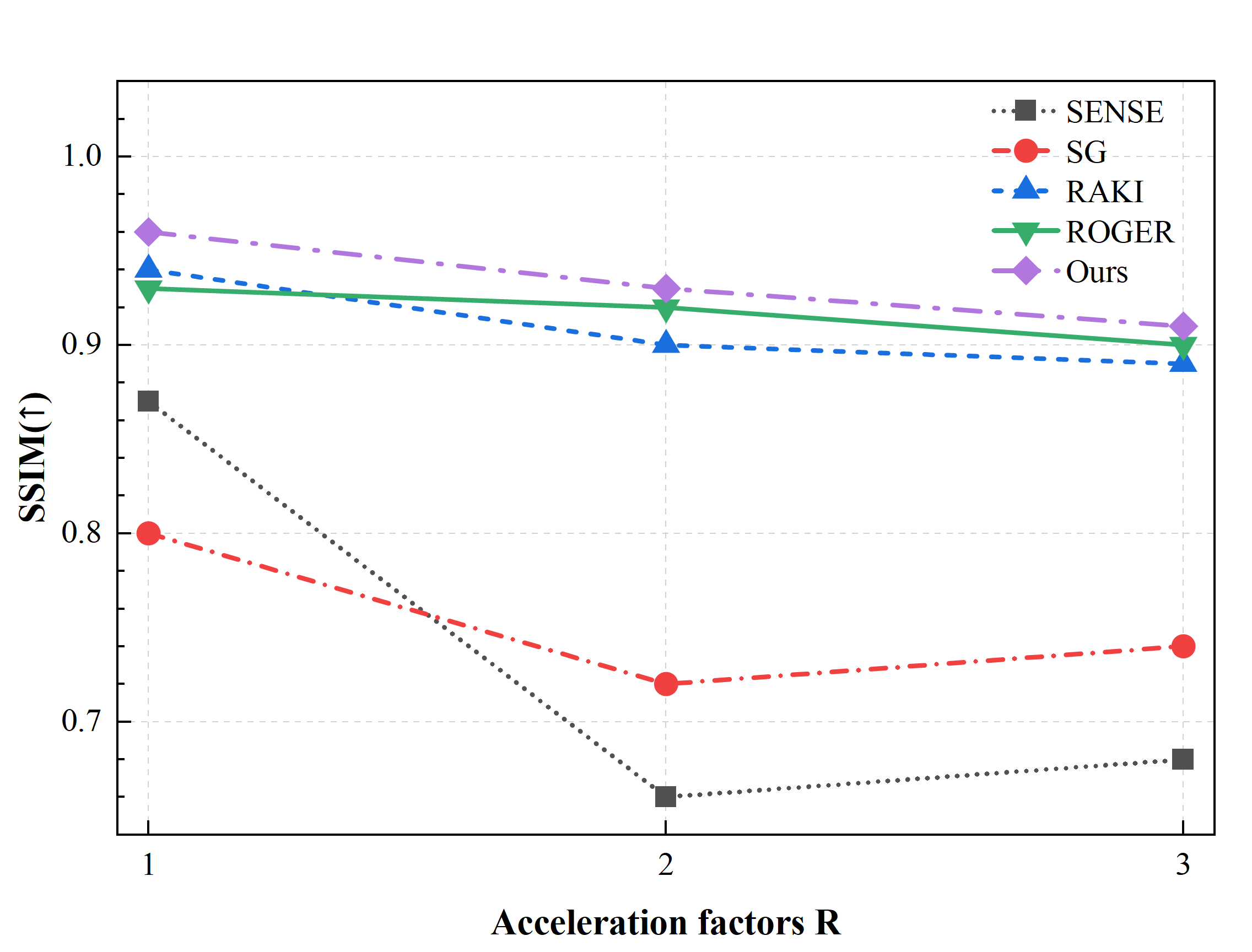}
  \vspace{-2mm}
  \caption{Quantitative comparison on fastMRI under MB3 with varying in-plane acceleration factors. Left: PSNR. Right: SSIM.NMSE trends are consistent and omitted for clarity.}
  \label{fig:fastmri_mb3_quant}
  \vspace{-3mm}
\end{figure*}

\subsubsection{Quantitative Results on fastMRI}
\label{sec:results_quant}

Fig.~\ref{fig:fastmri_mb3_quant} summarizes quantitative results on the fastMRI brain dataset
under MB3 with different in-plane acceleration factors.
Across all settings, the proposed method achieves the highest PSNR and SSIM among
all compared approaches.

This consistent improvement indicates that the proposed framework more effectively
handles the joint ill-posedness caused by SMS folding and in-plane undersampling.
We attribute this behavior to the operator-induced structured degradation inference trajectory,
which explicitly models SMS superposition and in-plane sampling within the generative
process, rather than relying on stochastic noise corruption.

\begin{figure*}[!t]
  \centering
  \includegraphics[width=0.45\textwidth]{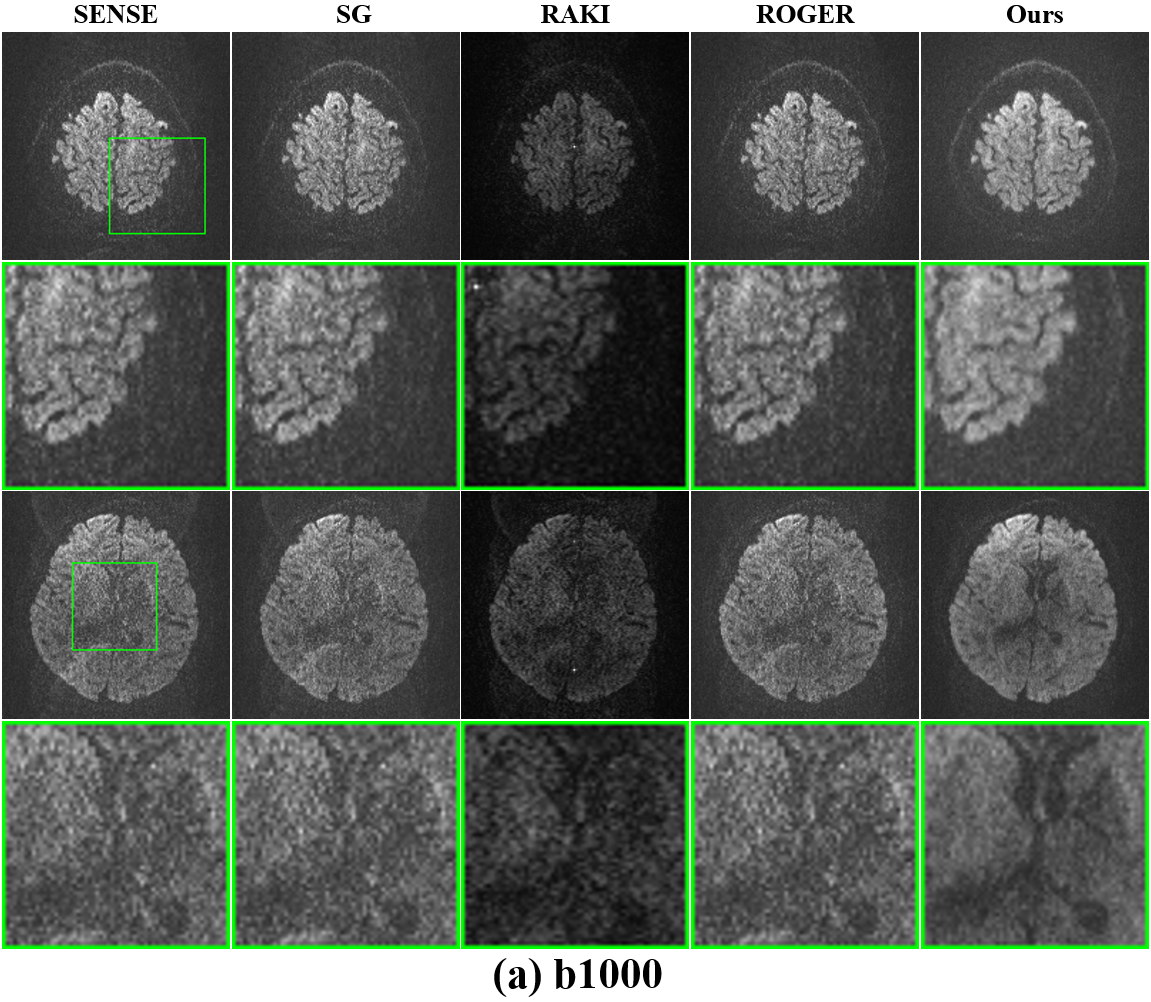}
  \includegraphics[width=0.45\textwidth]{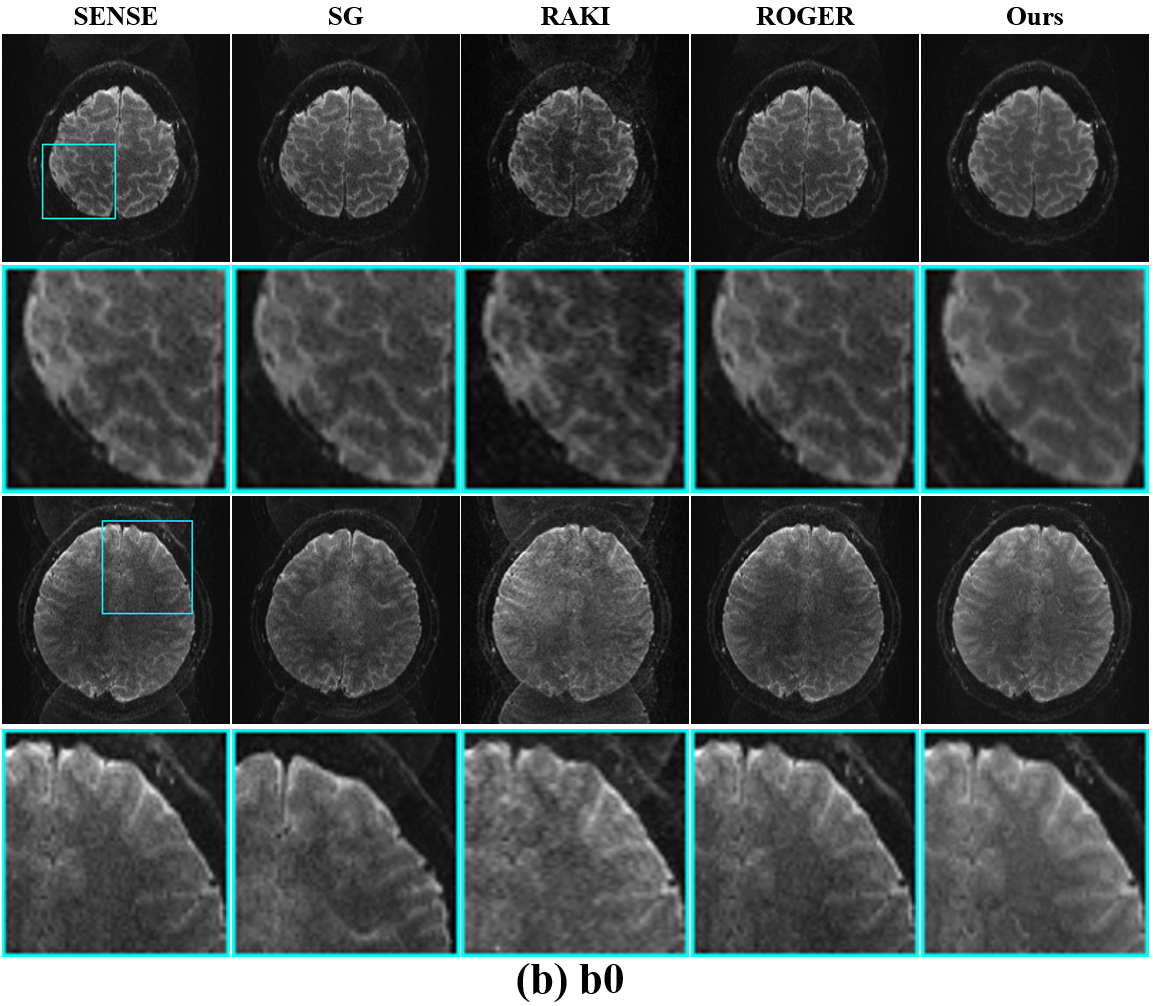}
  \vspace{-2mm}
  \caption{Prospective TJU 3.0T DWI results without ground truth.
(a) $b\!=\!1000$ MB2R2, where the model is trained using MB1 reconstructions.
(b) Cross-contrast generalization to $b\!=\!0$ MB2R2, where the model is trained on $b\!=\!1000$ and tested on $b\!=\!0$.
Two representative slices and zoomed-in ROIs are shown in each panel.}
  \label{fig:tju_b1000_and_b0}
  \vspace{-3mm}
\end{figure*}

\subsection{Prospective Evaluation}
\label{sec:prospective}

\subsubsection{Prospective Evaluation under Real Acquisition}
\label{sec:prospective_b1000}

The proposed method evaluate the proposed method on prospective DWI acquired with MB2R2 at $b=1000~\mathrm{s/mm^2}$.
Fully sampled slice-wise references are unavailable, so we focus on qualitative comparisons.
Fig.~\ref{fig:tju_b1000_and_b0}(a) shows two representative slices and corresponding zoomed ROIs for prospective MB2R2 DWI at $b=1000~\mathrm{s/mm^2}$.

SENSE and Slice-GRAPPA exhibit noticeable noise amplification and granular texture in the parenchyma.
These artifacts become more evident in the zoomed ROIs, where local tissue patterns appear mottled and less coherent.
RAKI produces substantially attenuated signals in both full-FOV views and ROIs, leading to reduced visual contrast and suppressed fine structures.
ROGER yields smoother appearances but still shows patchy residuals and blurred local variations inside the ROIs.
In contrast, OCDI produces more coherent local texture and clearer tissue boundaries in the ROIs. This advantage is aligned with the proposed operator-guided reverse updates, where the network explicitly predicts operator-induced degradations instead of denoising assumed Gaussian perturbations.

As shown in Fig.~\ref{fig:tju_b1000_and_b0}(b), OCDI yields clearer tissue boundaries and more coherent parenchymal texture than the compared methods.
It also reduces structured residual artifacts that are commonly observed in EPI-based DWI reconstructions.
These observations suggest that the operator-aligned inference remains effective under real acquisition conditions.

\subsubsection{Generalization Across Diffusion Contrasts}
\label{sec:prospective_b0}

Cross-contrast generalization is further tested without retraining.
Specifically, we apply the same model trained on $b=1000~\mathrm{s/mm^2}$ data to prospective MB2R2 data at $b=0~\mathrm{s/mm^2}$.
Fig.~\ref{fig:tju_b1000_and_b0}(b) shows that the contrast shift changes the global intensity distribution and local edge appearance, which challenges methods that rely on contrast-specific priors.

SENSE and Slice-GRAPPA show blurred cortical boundaries and reduced sharpness of sulcal patterns in the zoomed ROIs.
RAKI presents visible intensity non-uniformity and reduced local definition in the ROIs, and it may also introduce overly smoothed appearances.
ROGER improves smoothness but still exhibits loss of fine anatomical details, especially along curved cortical structures.
OCDI maintains clearer cortical boundaries and more consistent local appearance across the ROIs.
These observations suggest that the operator-aligned inference can provide improved robustness under diffusion-contrast shift.

\begin{table}[t]
\centering
\caption{Component-wise ablation on fastMRI under MB3R1. Best results are in \textbf{bold}.}
\label{tab:ablation_mb3r1_main}
\vspace{-1mm}
\setlength{\tabcolsep}{7pt}
\renewcommand{\arraystretch}{1.15}
\begin{tabular}{lccc}
\toprule
Variant & PSNR$\uparrow$ & SSIM$\uparrow$ & NMSE$\downarrow$ \\
\midrule
Single-stream & 40.19 & 0.958 & $3.8\times 10^{-3}$ \\
No cross-attention & 37.92 & 0.949 & $6.6\times 10^{-3}$ \\
OCDI-Net & \textbf{41.88} & \textbf{0.966} & $\boldsymbol{2.5\times 10^{-3}}$ \\
\bottomrule
\end{tabular}
\vspace{-2mm}
\end{table}

\subsection{Ablation Study}
\label{sec:ablation}

All ablation experiments are conducted on fastMRI T2 simulated SMS data under the MB3R1 setting.
We deliberately choose $\mathrm{MB}=3$ and $R=1$ to focus on the slice separation capability of the proposed model.
Under $R=1$, no in-plane undersampling is present and the reconstruction reduces to a single-stage SMS unaliasing problem.
This setting avoids error propagation across chained stages and eliminates confounding effects introduced by in-plane completion,
thereby providing a cleaner and more diagnostic testbed for evaluating how effectively the model disentangles and suppresses structured inter-slice interference.

Training and evaluation protocols follow Sec.~\ref{sec:exp_setup}.
PSNR, SSIM, and NMSE are computed on RSS magnitude images using the same normalization and evaluation pipeline as in the main experiments.

Table~\ref{tab:ablation_mb3r1_main} reports the main architectural ablation results.
Replacing the dual-stream OCDI-Net with a single-stream backbone consistently degrades reconstruction quality,
indicating that explicitly separating target slice content from interference helps reduce feature entanglement under multiband superposition.
Moreover, disabling cross-stream interaction while keeping the dual-stream structure leads to a noticeable performance drop,
highlighting that controlled information exchange between the content and interference streams is essential for effective suppression of structured slice leakage.

We further examine the role of attention in cross-stream interaction, as summarized in Table~\ref{tab:ablation_attention}.
When attention-based interaction is removed and replaced by local convolutional mixing, reconstruction fidelity deteriorates.
Reducing the attention capacity yields intermediate performance, while the full attention configuration achieves the best results.
These observations suggest that expressive cross-stream interaction is beneficial for modeling globally structured aliasing patterns induced by SMS acquisition.

\begin{table}[t]
\centering
\caption{Ablation on the attention capacity within cross-stream interaction under MB3R1. Best results are in \textbf{bold}.}
\label{tab:ablation_attention}
\vspace{-1mm}
\setlength{\tabcolsep}{7pt}
\renewcommand{\arraystretch}{1.15}
\begin{tabular}{lccc}
\toprule
Variant & PSNR$\uparrow$ & SSIM$\uparrow$ & NMSE$\downarrow$ \\
\midrule
No attention & 39.64 & 0.955 & $4.2\times 10^{-3}$ \\
Reduced attention & 40.87 & 0.963 & $3.3\times 10^{-3}$ \\
Global attention & \textbf{41.88} & \textbf{0.966} & $\boldsymbol{2.5\times 10^{-3}}$ \\
\bottomrule
\end{tabular}
\vspace{-2mm}
\end{table}

\FloatBarrier


\section{Discussion}
\label{sec:discussion}

Our study suggests that an operator-aligned and deterministic degradation chain can better reflect SMS acquisition than a Gaussian-noise diffusion process.
This design keeps each intermediate state tied to the assumed acquisition operators, so the reverse updates target structured inter-slice interference and missing-data corruption.
The quantitative gains on fastMRI and the reduced slice-leakage patterns in visual comparisons are consistent with this interpretation.
This view also clarifies the role of the network, which predicts structured degradations that are meaningful under the assumed forward model.

We further analyze ablations and sensitivity experiments to understand which design choices contribute most to the observed gains.
The results indicate that separating target content from interference reduces feature entanglement, and controlled cross-stream interaction improves suppression of coherent leakage.

At the same time, the method depends on modeling assumptions that may not always hold in practice.
The reverse process relies on accurate CAIPI modulation and a correct in-plane sampling mask, so operator mismatch can leave structured residual artifacts.
In addition, the two-stage pipeline may propagate errors, because Stage-U is initialized from Stage-M and may inherit its mistakes under very limited support.

Several limitations remain and motivate future work.
More prospective validation across scanners, coil configurations, and CAIPI patterns is needed to quantify robustness under protocol variations.
Future work may extend the operator set to cover richer physical effects and may adapt the framework to non-Cartesian sampling when needed.
Controlled mismatch studies that perturb CAIPI or sampling patterns can also provide a direct stress test of operator sensitivity.

\section{Conclusion}
\label{sec:conclusion}

This paper presented an operator-guided k-space reconstruction framework for accelerated SMS MRI under combined multiband excitation and in-plane undersampling.
The method constructs a deterministic degradation trajectory governed by the acquisition operators and performs deterministic inversion of structured slice mixing and missing data.
Under this formulation, we introduced OCDI-Net, which separates target-slice content from inter-slice interference and enables controlled information exchange for effective suppression.
Experiments on retrospectively simulated fastMRI and prospective DWI acquisitions demonstrated improved reconstruction fidelity and reduced slice leakage compared with conventional and learning-based baselines.
These results suggest that aligning the inference trajectory with the acquisition operators is a practical way to improve robustness under stronger acceleration.

\end{document}